\documentclass[10pt,twocolumn,letterpaper]{article}

\usepackage{cvpr}      
\usepackage{float}
\usepackage{times}
\usepackage{latexsym}
\usepackage{graphicx}
\usepackage{amsmath}
\usepackage{amsfonts}
\usepackage{enumitem}
\usepackage{bm}
\usepackage{bbding}
\usepackage{multirow}
\usepackage{booktabs}
\usepackage[T1]{fontenc}
\usepackage[utf8]{inputenc}
\usepackage{microtype}
\usepackage{makecell}
\usepackage{adjustbox}
\usepackage{subcaption}
\usepackage{multirow}
\usepackage{makecell}
\usepackage{ulem}
\usepackage{amssymb}

\usepackage[dvipsnames]{xcolor}

\definecolor{cvprblue}{rgb}{0.21,0.49,0.74}
\usepackage[pagebackref,breaklinks,colorlinks,citecolor=cvprblue]{hyperref}

\def\paperID{2992} 
\def\confName{CVPR}
\def\confYear{2024}

\title{Efficient Temporal Sentence Grounding in Videos with Multi-Teacher Knowledge Distillation}

\author{Renjie Liang\\
National University Singapore\\
{\tt\small liangrj5@gmail.com}
\and
Yiming Yang \\
National University Singapore\\
{\tt\small e0920761@u.nus.edu}
\and
Hui Lu \\
Nanyang Technological University\\
{\tt\small hui007@ntu.edu.sg}
\and
Li Li \\
National University Singapore\\
{\tt\small lili02@u.nus.edu}
}

\begin{document}

\maketitle

\begin{abstract}
Temporal Sentence Grounding in Videos (TSGV) aims to retrieve the event timestamps described by the natural language query from untrimmed videos. This paper discusses the challenge of achieving efficient computation in TSGV models while maintaining high performance. Most existing approaches exquisitely design complex architectures to improve accuracy, suffering from inefficiency and heaviness. Previous attempts to address this issue have primarily concentrated on feature fusion layers. To tackle this problem, we propose a novel efficient multi-teacher model (EMTM) based on knowledge distillation to transfer diverse knowledge from multiple networks. 
Specifically, We first unify different outputs of the different models. Next, a Knowledge Aggregation Unit (KAU) is built to acquire high-quality integrated soft labels from multiple teachers. Additionally, we propose a Shared Encoder strategy to enhance the learning of shallow layers. Extensive experimental results on three popular TSGV benchmarks demonstrate that our method is both effective and efficient. Our code is available at https://github.com/renjie-liang/EMET.

\end{abstract}



\vspace{-5mm}
\section{Introduction}
\vspace{-2mm}

\begin{figure}[htbp]

    \centering
    \begin{subfigure}{\columnwidth}
    \centering
        \includegraphics[width=0.7\columnwidth]{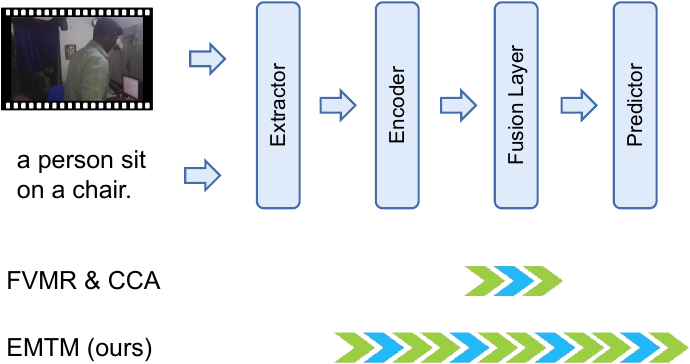}
    \caption{}
    \label{fig:accalerate}
    \end{subfigure}

    \centering
    \begin{subfigure}{\columnwidth}
    \centering
        \includegraphics[width=0.7\columnwidth]{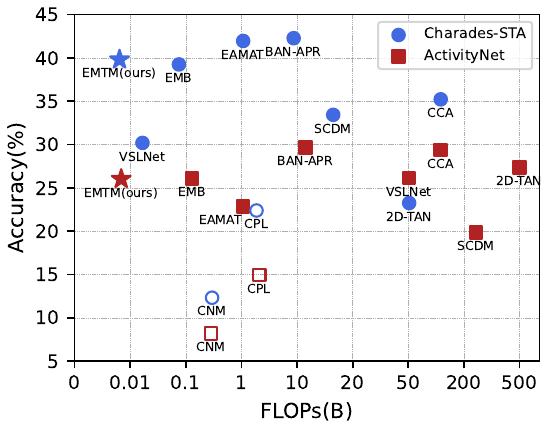}
        
    \caption{}

    \label{fig:relation_speed_accurcy}
    \end{subfigure}

    \caption{
    (a) The previous FVMR and CCA methods expedite the Fusion Layer, and our EMTM accelerates the Encoder, Fusion Layer, and Predictor.
    (b) FLOPs and accuracy plot of state-of-the-art TSGV approaches on Chradas-STA and ActivityNet. We report R1@0.7 for the two datasets. Our proposed EMTM achieves the best accuracy-speed balance among all the competitors. }

    \label{fig:introduction}
    \vspace{-3mm}
\end{figure}

Temporal Sentence Grounding in Videos (TSGV), which aims to ground a temporal segment in an untrimmed video with a natural language query, has drawn widespread attention over the past few years \cite{10075491}. There is a clear trend that top-performing models are becoming larger with numerous parameters. 
Additionally, the recent work shows that accuracy in TSGV tasks has reached a bottleneck period, while the combination of complex networks and multiple structures is becoming more prevalent to further improve the ability of the model,
which will cause an expansion of the model size. However, the heavy resource cost required by the approaches restricts their applications.

In order to improve efficiency, FMVR \cite{gao2021fast} and CCA \cite{wu2022learning} are proposed to construct fast TSGV models by reducing the fusion time.
However, they only decline the inferred time significantly, the whole network is still time-consuming, as depicted in Figure \ref{fig:accalerate}.
Another shortage is that FMVR and CCA are required to store the feature after the encoder with extra storage.

To tackle this challenge, we extend the accelerated span to the entire network, shown in Figure \ref{fig:accalerate}.
A natural approach is to reduce the complexity of the network, which can involve decreasing the hidden dimension, reducing the number of layers, and eliminating auxiliary losses. Nevertheless, all of these methods will lead to a decrease in performance to some extent.
One promising technique is knowledge distillation \cite{hinton2015distilling} to mitigate the decrease in performance and maintain high levels of accuracy when lighting the network. 

Initially, knowledge distillation employed a single teacher, but as technology advanced, multiple teachers have been deemed beneficial for imparting greater knowledge\cite{fukuda2017efficient}, as extensively corroborated in other domains\cite{wang2021knowledge}.
Multi-teacher strategy implies that there is a more diverse range of dark knowledge to be learned, with the optimal knowledge being more likely to be present \cite{10.1145/3097983.3098135}. 
Thus far, multiple-teacher knowledge distillation has not been studied and exploited for the TSGV task.

An immediate problem is that different models will produce heterogeneous output, e.g., candidates for proposed methods, or probability distribution for proposal-free methods. A question is how to identify optimal knowledge from multiple teachers.
In addition, knowledge was hardly backpropagted to the front layers from the soft label in the last layers \cite{romero2014fitnets}, meaning that the front part of the student model usually hardly enjoys the benefit of teachers' knowledge. 
Until now, here are three issues we need to deal with: i) how to unify knowledge from the different models, ii) how to select the optimal knowledge and assign weights among these teachers, and iii) how the front layers of the student benefit from the teachers.

Several challenges arise: Firstly, different models may yield heterogeneous outputs such as varying candidate spans or probability distributions. 
Secondly, backpropagating knowledge is difficult from the last to the front layers, limiting the student shallow layer to learn from teacher models \cite{romero2014fitnets}.


For the first challenge, we unify the heterogeneity of model outputs by converting them into a unified 1D probability distribution. This enables us to seamlessly integrate the knowledge during the model training.
The 1D distribution, derived from the span-based method in the proposal-free catalog, offers a speed advantage over proposal-based methods.

To integrate knowledge from various models, we have developed a Knowledge Aggregation Unit (KAU). This unit utilizes multi-scale information \cite{li2021online} to derive a higher-quality target distribution, moving beyond mere averaging of probabilities. The KAU adaptively assigns weights to different teacher models. This approach overcomes the limitations of manually tuning weights, a sensitive hyperparameter in multi-teacher distillation  \cite{liu2022meta}.

Regarding the second challenge, we implemented a shared layer strategy to facilitate the transfer of shallow knowledge from the teacher to the student model. This involves co-training a teacher model with our student model, sharing encoder layers and aligning hidden states. Such an arrangement ensures comprehensive and global knowledge acquisition by the student model.

During inference, we only exploit the student model to perform inference, which does not add computational overhead.
To sum up, this paper's primary contributions can be distilled into three main points, which are outlined below:

\begin{itemize}[leftmargin=*]
    \item 
    We propose a multi-teacher knowledge distillation framework for the TSGV task. This approach substantially reduces the time consumed and significantly decreases the number of parameters, while still maintaining high levels of accuracy.

    \item  
    To enable the whole student to benefit from various teacher models, we unify the knowledge from different models and use the KAU module to adaptively integrate to a single soft label. Additionally, a shared encoder strategy is utilized to share knowledge from the teacher model in front layers.

    \item 
    Extensive experimental results on three popular TSGV benchmarks demonstrate that our proposed method performs superior to the state-of-the-art methods and has the highest speed and minimal parameters and computation.
\end{itemize}

\section{Related Work}
\vspace{-2mm}

Given an untrimmed video, temporal sentence grounding in videos (TSGV) is to retrieve a video segment according to a query, which is also known as Video Moment Retrieval (VMR). 
Existing solutions to video grounding are roughly categorized into proposal-based and proposal-free frameworks. We also introduce some works on fast video temporal grounding as follows.

\subsection{Proposal-based Methods}
\vspace{-2mm}
The majority of proposal-based approaches rely on some carefully thought-out dense sample strategies, which gather a set of video segments as candidate proposals and rank them in accordance with the scores obtained for the similarity between the proposals and the query to choose the most compatible pairs. 
\citet{zhang2020learning} convert visual features into a 2D temporal map and encode the query in sentence-level representation, which is the first solution to model proposals with a 2D temporal map (2D-TAN). 
BAN-APR \cite{dong2022boundary} utilize a boundary-aware feature enhancement module to enhance the proposal feature with its boundary information by imposing a new temporal difference loss. Currently, most proposal-based methods are time-consuming due to the large number of proposal-query interactions.

\subsection{Proposal-free Methods}
\vspace{-2mm}

Actually, the caliber of the sampled proposals has a significant impact on the impressive performance obtained by proposal-based methods. To avoid incurring the additional computational costs associated with the production of proposal features, proposal-free approaches directly regress or forecast the beginning and end times of the target moment.
VSLNet \citet{zhang2020span} exploits context-query attention modified from QANet \citet{yu2018fast} to perform fine-grained multimodal interaction. Then a conditioned span predictor computes the probabilities of the start/end boundaries of the target moment. 
SeqPAN \cite{zhang2021parallel} design a self-guided parallel attention module to effectively capture self-modal contexts and cross-modal attentive information between video and text inspired by sequence labeling tasks in natural language processing. 
\citet{yang2022entity} propose Entity-Aware and Motion-Aware Transformers (EAMAT) that progressively localize actions in videos by first coarsely locating clips with entity queries and then finely predicting exact boundaries in a shrunken temporal region with motion queries. 
Nevertheless, with the improvement of performance, huge and complex architectures inevitably result in higher computational cost during inference phase.

\subsection{Fast Video Temporal Grounding}
\vspace{-2mm}
Recently, fast video temporal grounding has been proposed for more practical applications.
According to \cite{gao2021fast}, the standard TSGV pipeline can be divided into three components. The visual encoder and the text encoder are proved to have little influence in model testing due to the features pre-extracted and stored at the beginning of the test, and cross-modal interaction is the key to reducing the test time. Thus, a fine-grained semantic distillation framework is utilized to leverage semantic information for improving performance. Besides, \citet{wu2022learning} utilize commonsense knowledge to obtain bridged visual and text representations, promoting each other in common space learning.
However, based on our previous analysis, the inferred time proposed by \cite{gao2021fast} is only a part of the entire prediction processing. The processing from inputting video features to predicting timestamps is still time-consuming. 

\begin{figure*}
    \centering
    \includegraphics[width=\linewidth,height=0.5\linewidth]{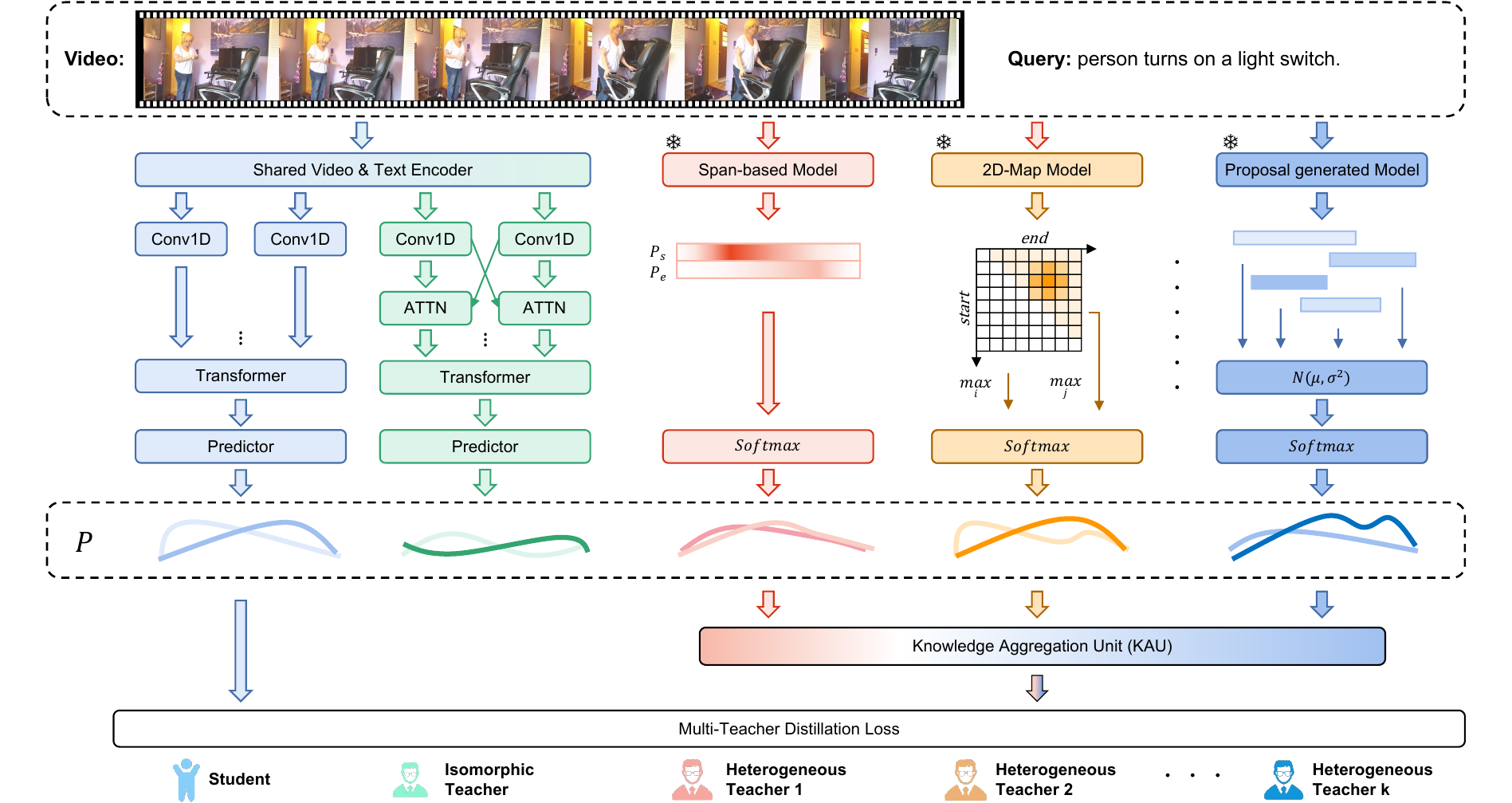}
    \caption{An overview of the proposed framework. EMTM mainly consists of three components: the student model, the shared encoder, and the KAU. The shared encoder is utilized to align their hidden states. The label teacher model outputs are unified into 1D probability distribution as shown on the right. Then it is adopted to adaptively determine the importance weights of different teachers with respect to a specific instance based on both the teacher and student representation. During the inference stage, only the student model is adopted for fast TSGV. 
    }
    \label{fig:pipeline}
    \vspace{-5mm}
\end{figure*}

\section{Methodology}
\vspace{-2mm}

In this section, we first give a brief task definition of TSGV in Section \ref{sec:problem_formulation}. In the following, heterogeneous knowledge unification is presented as a prerequisite in Section \ref{sec:unify_knowledge}. Then we introduce the student network (Section \ref{sec:student}), shared encoder strategy (Section \ref{sec:encoder}) and knowledge aggregation unit (Section \ref{sec:KAU}) as shown in \ref{fig:pipeline}.

Finally, the training and inference processes are presented in section \ref{sec:training_inference}, as well as the loss settings.

\subsection{Problem Formulation}\label{sec:problem_formulation}
\vspace{-2mm}
Given an untrimmed video $V=[f_t]_{t=1}^{T}$ and the language query $Q=[q_j]_{j=1}^{m}$, where $T$ and $m$ are the numbers of frames and words, respectively. The start and end times of the ground truth moment are indicated by $\tau^{s}$ and $\tau^{e}$,  $1\leq\tau_s<\tau_e\leq T$.
Mathematically, TSGV is to retrieve the target moment starting from $\tau_s$ and ending at $\tau_e$ by giving a video $V$ and query $Q$,  i.e., $\mathcal{F}_{TSGV}:(V, Q)\mapsto (\tau_s,\tau_e)$.

\subsection{General Scheme}\label{sec:general_scheme}

\vspace{-2mm}
\subsubsection{Heterogeneous Knowledge Unification} \label{sec:unify_knowledge}
\vspace{-2mm}

Compared to the proposal-based method, the span-based method doesn't need to generate redundant proposals, which is an inherent advantage in terms of efficiency. Meanwhile, 1D distribution carries more knowledge than the regression-based method. 
Hence we unify various heterogeneous outputs into 1D probability distribution and develop our network based on the span-based method, as shown in Figure \ref{fig:pipeline}.  The outputs of the span-based method are the 1D probability distributions of start and end moments, denoted as $P_s, P_e \in \mathbb{R}^n$. To keep concise, we adopt $P \in \mathbb{R}^{2n}$ without subscripts to express stacked probability for the start and end moments.

We simply adopt the softmax function to the outputs of the span-based methods and obtain probability distributions.
\begin{equation}
    \begin{aligned}
       P_s = Softmax(P'_s) \quad
       P_e = Softmax(P'_e)
    \end{aligned}
\end{equation}

2D-map anchor-based method is a common branch of the proposal-base method, such as \cite{zhang2020learning}, \cite{dong2022boundary}. A 2D map $S =[s_{i,j}] \in \mathbb{R}^{n \times n}$ is generated
to model temporal relations between proposal candidates, on which one dimension indicates the start moment and the other indicates the end moment. We calculate the max scores of $S$ by row/column as start/end distributions.

\begin{equation}
    \begin{aligned}
    P_s =& Softmax(\max\limits_{j} s_{i,j})  \\
    P_e =& Softmax(\max\limits_{i} s_{i,j})
    \end{aligned}
\end{equation}

As for the regression-based method, we can get a time pair $(t_s, t_e)$ after computation. Then the Gaussian distribution is leveraged to simulate the probability distribution of the start/end moments as follows:

\begin{equation}
    \begin{aligned}
       P_s =& Softmax({N(t_s, \sigma^2)})  \\
       P_e =& Softmax({N(t_e, \sigma^2)})
    \end{aligned}
\end{equation}

The proposal-generated method will generate a triple candidate list $S'=(t_s^i, t_e^i, r^i) \in \mathbb{R}^{3 \times k}$, where $k$ is the number of proposal candidates. Similarly, we use the Gaussian distribution to generate the probability distribution of the start/end moment for each candidate. Then we put different weights on various candidates and accumulate them:

\begin{equation}
    \begin{aligned}
       P_s =& Softmax(\sum_i r^i {N(t_s^i, \sigma^2)})  \\
       P_e =& Softmax(\sum_i r^i {N(t_e^i, \sigma^2)})
    \end{aligned}
\end{equation}
where $\sigma^2$ is the variance of Gaussian distribution $N$.

\vspace{-2mm}

\subsubsection{Student Network}\label{sec:student}
\vspace{-2mm}

The design of the student network emphasizes efficient processing. The video feature, represented as $\bm{V} \in \mathbb{R}^{n \times d_v}$, utilizes the I3D framework, as proposed by Carreira et al. \cite{carreira2017quo} and further refined by Zhang \cite{zhang2021parallel}.
In parallel, the query feature, denoted by $\bm{Q} \in \mathbb{R}^{m \times d_q}$, is initialized using GloVe embeddings, with $n$ indicating the length of the features. 
Both the video and query features undergo projection and encoding processes, which align their dimensions to facilitate uniformity and interoperability in subsequent computational stages.

\begin{equation}
\begin{aligned}
    \bm{V'} & = \mathtt{VisualEncoder}(\bm{V}) 
    \\
    \bm{Q'} & = \mathtt{QueryEncoder}(\bm{Q})
\end{aligned}
\label{eq:projector}
\end{equation}

Subsequently, a lightweight transformer is employed to fuse the video and query features. This approach, in contrast to the direct dot multiplication method referenced in Gao et al. \cite{gao2021fast}, significantly enhances performance without imposing excessive computational load on the model.

\begin{equation}
\bm{V}^{qv}=\mathtt{Transformer}( \bm{V'},  \bm{Q'})
\end{equation}

Finally, a predictor is utilized to generate the logits corresponding to the start and end points. Then multiply $\bm{P_{s}}$ and $\bm{P_{e}}$ to form a matrix, within which the highest value is identified. The row and column indices of this peak value correspond to the predicted start and end indices, respectively.

\begin{equation}
    \begin{aligned} 
    (\bm{P_{s}},\bm{P_{e}}) = \mathtt{Predictor} (\bm{V^{qv}}) \in\mathbb{R}^{n}
    \end{aligned} 
\end{equation}

\vspace{-2mm}
\subsubsection{Teacher Network}\label{sec:teacher}
\vspace{-2mm}

The teacher network is architected with a focus on performance optimization, in contrast to the student network which is streamlined for efficiency. The specific differences between the two networks are comprehensively itemized in Table \ref{tabel:teacher_student}. Cross-modal fusion layers have been incorporated to augment the interactivity among disparate modalities within the teacher model.  
Following the Encoder stage, a group of transformations is employed to facilitate the amalgamation of query and video features. Subsequently, these cross-modal features are fed into the fusion layer.

\begin{equation}
\begin{aligned}
    \bm{V'} & = \mathtt{Transformer}(\bm{V'},\bm{Q'} ) 
    \\
    \bm{Q'} & = \mathtt{Transformer}(\bm{Q'}, \bm{V'} )
\end{aligned}
\label{eq:layer:cross}
\end{equation}

\begin{table}[htbp]
\footnotesize
\centering
\begin{adjustbox}{width=0.9\linewidth}
\centering

\begin{tabular}{l|l|c|c}
\toprule

Module                     & Layer               & Teacher & Student \\ \hline 
\multirow{2}{*}{Projector} & CONV1D\_query       & 1       & 1       \\ \cline{2-4} 
                           & CONV1D\_video       & 4       & 4       \\ \hline
\multirow{3}{*}{Fusion Layer}  & Transformer\_query  & 4       & 0       \\ \cline{2-4} 
                           & Transformer\_video  & 4       & 0       \\ \cline{2-4} 
                           & Transformer\_fusion & 1       & 1       \\ \hline
\multirow{2}{*}{Predictor} & CONV1D\_start       & 6       & 3       \\ \cline{2-4} 
                           & CONV1D\_end         & 6       & 3       \\ 
\bottomrule
\end{tabular}
\end{adjustbox}
\caption{Comparative overview of key distinctions between the teacher and student networks, excluding minor details for clarity}
\label{tabel:teacher_student}
\vspace{-4mm}
\end{table}

\begin{table*}[htbp]
\footnotesize
\begin{adjustbox}{width=\textwidth}
\renewcommand*{\arraystretch}{1.2}
\setlength{\tabcolsep}{1pt}

\label{table1}
\begin{tabular}{c|c|rrrr|rrrr|rrrr}

\toprule
\multirow{2}{*}{Method} & \multirow{2}{*}{Year} & \multicolumn{4}{c|}{Charades-STA}            & \multicolumn{4}{c|}{ActivityNet}             & \multicolumn{4}{c}{TACoS}  \\ \cline{3-14} 
      &     & FLOPS (B) & Params (M) & Times (ms) & sumACC & FLOPS (B) & Params (M) & Times (ms) & sumACC & FLOPS (B) & Params (M) & Times (ms) & sumACC \\ \hline
SCDM  & 2019   & 16.5000   & 12.8800    & -          & 87.87  & 260.2300  & 15.6500    & -          & 56.61  & 260.2300  & 15.6500    & -          & -      \\
2D-TAN   & 2020   & 52.2616   & 69.0606    & 13.3425    & 66.05  & 1067.9000 & 82.4400    & 77.9903    & 71.43  & 1067.9000 & 82.4400    & 77.9903    & *36.82 \\
VSLNet   & 2020   & 0.0300    & \underline{0.7828}     & 8.0020     & 77.50  & 0.0521   & \underline{0.8005}     & 8.9893     & 69.38  & 0.0630   & 0.8005     & 8.9893     & 44.30  \\
SeqPAN   & 2021   & \underline{0.0209}    & 1.1863     &  10.5168          & 102.20 &  \underline{0.0214}	& 1.2143	 & 13.7138       & 73.87  &  0.0218	& 1.2359 &	23.3025        & \textbf{67.71}  \\
EMB   & 2022   & 0.0885    & 2.2168     & 22.3900    & 97.58  &      0.2033 &	6.1515	& 25.0871	&  70.88  &      0.2817 &	2.2172 &	23.6349	& 60.36  \\
EAMAT & 2022   & 1.2881    & 94.1215    & 56.1753    & \underline{103.65} & 4.1545    & 93.0637    & 125.7822   & 60.94  & 4.1545    & 93.0637    & 125.7822   & \underline{64.98}  \\
BAN-APR  & 2022   & 9.4527    & 34.6491    & 19.9767    & \textbf{105.96} & 25.4688   & 45.6714    & 44.8587    & \textbf{77.79}  & 25.4688   & 45.6714    & 44.8587    & *52.10 \\ \hline
CPL   & 2022   & 3.4444    & 5.3757     & 26.8451    & 71.63  & 3.8929    & 7.0115     & 26.4423    & 49.14  & -         & -          & -          & -      \\
CNM   & 2022   & 0.5260    & 5.3711     & \underline{5.4482}     & 50.10  & 0.5063    & 7.0074     & \underline{4.8629}     & *48.96 & -         & -          & -          & -      \\ \hline
FVMR   & 2021   & -  & -   & -    & 88.75  & -  & -    & -    & 71.85 & -  & -    & -    & -  \\ 
CCA   & 2022   & 137.2984  & 79.7671    & 26.9734    & 89.41  & 151.1023  & 22.5709    & 31.5400    & \underline{*75.95} & 151.1023  & 22.5709    & 31.5400    & 50.90  \\ \hline
EMTM (Ours)  &     & \textbf{0.0081}	& \textbf{0.6569}	& \textbf{4.7998}    &  92.80   &   \textbf{0.0084}	 & \textbf{0.6848}	& \textbf{3.5431}	& 70.91        &  \textbf{ 0.0087}	& \textbf{0.7065} &	\textbf{4.5737}   &58.24    \\ \bottomrule
\end{tabular}
\end{adjustbox}
\caption{Efficiency analysis on Charades-STA, ActivityNet, and TACoS. \textbf{sumACC} is the sum of R1@0.5 and R1@0.7. All the data is measured with strict adherence to the source code in the same environment. * denotes the accuracy we reproduce.}
\label{table:efficiency}
\vspace{-5mm}
\end{table*}

\vspace{-3mm}
\subsubsection{Knowledge Aggregation Unit}\label{sec:KAU}
\vspace{-2mm}

\begin{figure}[htbp]
    \centering
    \includegraphics[width=\linewidth]{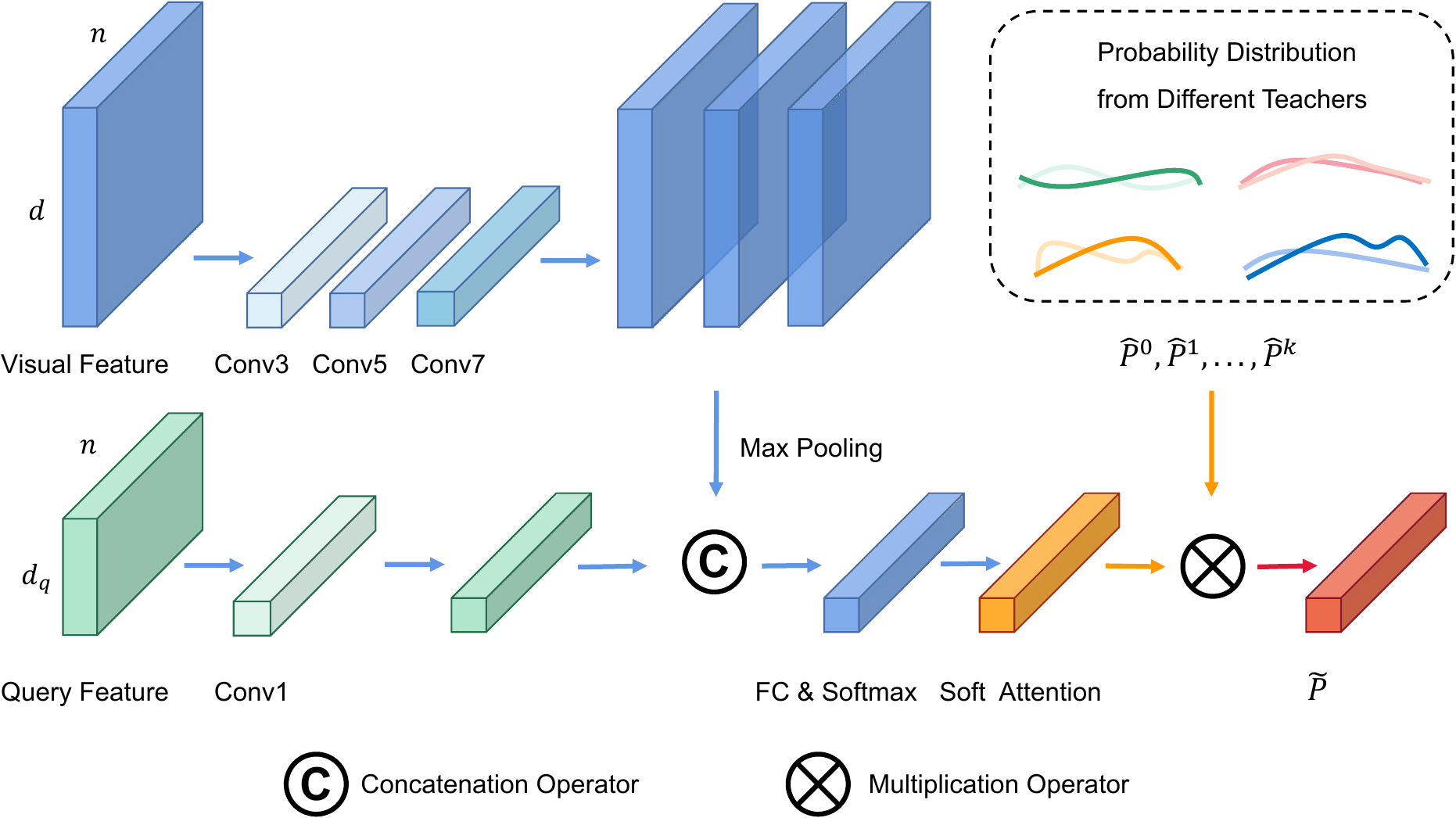}

    \caption{Illustration of Knowledge Aggregation Unit, which exploits the multi-scale information from various teachers to generate higher-quality knowledge. The final ensemble probability distribution $\widetilde{P}$ is obtained by the weighted sum from all individual branches.} 
    \label{fig:kau}
\vspace{-4mm}
\end{figure}

Our goal is to combine all the unified predictions from $b$ branches to establish a strong teacher distribution. 
Drawing inspiration from \cite{li2021online}, we developed the Knowledge Aggregation Unit (KAU). The KAU integrates parallel transformations with varying receptive fields, harnessing both local and global contexts to derive a more accurate target probability. The KAU's design is illustrated in Figure \ref{fig:kau}.

Considering saving more original information, we first take the video features $V'$  in the eq. (\ref{eq:projector}) as input and then add convolution layers. The convolution operation is conducted with a small kernel size of 3 initially and then consistently increased to 5 and 7.
Further, we incorporate the average pooling of query features $Q'$ in the eq. (\ref{eq:projector})  for richer representations. Then we concatenate all the splits and obtain the intermediate vector $v$, denoted as:
\begin{equation}
    \begin{aligned}
 v = [q_{avg}, g([v_{conv3}, v_{conv5}, v_{conv7}])]
    \end{aligned}
\end{equation}
where $q_{avg}$ denotes the result of $Q'$ after average pooling, g(·) denotes the global pooling function, $v_{conv3}$,  $v_{conv5}$,  and $v_{conv7}$ denote the results after the convolution layers with kernel size 3, 5, and 7, respectively.

Passing through a fully connected layer $FC$,  a channel-wise softmax operator is applied to obtain the soft attention $a$.
\begin{equation}
    \begin{aligned}
 a = Softmax(\mathtt{FC} (v)) \in \mathbb{R}^{2b \times n}
    \end{aligned}
\end{equation}
where the $b$ denote the number of teacher branch, $2b$ is because there are two probability distributions (i.e., start and end). 

Finally, we fuse prediction results from multiple branches via an element-wise summation to obtain the weighted ensemble probability.

\begin{equation}
    \begin{aligned}
 \widetilde{P} = \sum_{i=1}^{b}{a^i \otimes \hat{P}^{i} } \in \mathbb{R}^{2 \times n}
    \end{aligned}
\end{equation}
where $\widetilde{P}$ denotes the ensemble probability, $P^{i} \in \mathbb{R}^{2 \times n}$ means the start and end distribution from $i$-th teacher branch, and $\otimes$ refers to the channel-wise multiplication. Our experiments (see Section \ref{sec:main_components}) prove that the weights generated by KAU can achieve better distillation performance.

\vspace{-3mm}
\subsubsection{Shared Encoder Strategy}\label{sec:encoder}
\vspace{-2mm}

The backpropagation of knowledge from soft labels often provides limited benefit to the shallow layers, primarily due to the influence of non-linear activation functions and dropout mechanisms. However, the concept of feature invariance in these layers, in Zeiler et al.\cite{zeiler2014visualizing}, guides our approach. We propose the sharing of several shallow layers between the student and teacher networks. This collaborative training strategy enables the shallow layers in the student network to assimilate additional knowledge from the teacher network, enhancing their learning capacity.

Specifically, a student and a teacher share their text and query encoder, shown in Figure \ref{fig:pipeline}. The encoder consists of several conv1D in our network, which is lightweight and fast due to its inherent characteristics.
The $\mathtt{VisualEncoder}$,  $\mathtt{QueryEncoder}$ in eq. (\ref{eq:projector}) denote 
the shared layers in our network.

\subsection{Training and Inference} \label{sec:training_inference}
\vspace{-2mm}
\subsubsection{TSGV Loss}
\vspace{-2mm}

The overall training loss of our model is described as follows. For the student and the teacher, the hard loss (i.e. label loss) is used to optimize distributions of start/end boundaries.
\begin{equation}
    \begin{aligned}
 L^{st}_{loc} = f_{CE}(P^{st}, Y) \\
 L^{tc}_{loc} = f_{CE}(P^{tc}, Y)
    \end{aligned}
\end{equation}
where $f_{CE}$ is the cross-entropy function, and $Y$ is one-hot labels for the start and end boundaries of ground truth. Similarly, we encourage ensemble probability to get closer to ground truth distribution.
\begin{equation}
    \begin{aligned}
 L^{ens}_{loc} = f_{CE}(\widetilde{P}, Y) \\
    \end{aligned}
\end{equation}

As we discussed previously, the learned ensemble information serves as complementary cues to provide an enhanced supervisory signal to our student model. As a result, we introduce multiple distillation learning, which transfers the rich knowledge in the form of softened labels. The formulation is given by:
\begin{equation}
    \begin{aligned}
 L_{dis} = f_{KL}(softmax(P^{st}, t), softmax(\widetilde{P}, t) \\
    \end{aligned}
\end{equation}
where $f_{KL}$ represents the KL divergence. The $t$ is the temperature in knowledge distillation, which control the smoothness of the output distribution. 

Based on the above design, the overall objective for a training video-query pair is formulated as:
\begin{equation}
    \begin{aligned}
 L = L^{st}_{loc} + L^{tc}_{loc} + L^{ens}_{loc} + \alpha L_{dis} \\
    \end{aligned}
\end{equation}
where $\alpha$ is a balance term.

\vspace{-2mm}

\subsubsection{Inference}
\vspace{-2mm}

The teacher and student models will be collaboratively trained, while we only adopt the student model for TSGV during testing. The learned rich information serves as complementary cues to provide an enhanced supervisory signal to the TSGV model. 
Compared with  FMVR \cite{gao2021fast} and CCA \cite{wu2022learning}, we won't pre-calculate and store visual features.

\begin{table*}[]
\footnotesize
\begin{adjustbox}{width=\textwidth}
\begin{tabular}{c|rrrr|rrrr|rrrr}
\toprule
\multirow{2}{*}{Method} & \multicolumn{4}{c|}{Charades-STA} & \multicolumn{4}{c|}{ActivityNet} & \multicolumn{4}{c}{TACoS}   \\ \cline{2-13}
& \multicolumn{1}{c}{R1@0.3} & \multicolumn{1}{c}{R1@0.5} & \multicolumn{1}{c}{R1@0.7} & \multicolumn{1}{c|}{mIoU} & \multicolumn{1}{c}{R1@0.3} & \multicolumn{1}{c}{R1@0.5} & \multicolumn{1}{c}{R1@0.7} & \multicolumn{1}{c|}{mIoU} & \multicolumn{1}{c}{R1@0.3} & \multicolumn{1}{c}{R1@0.5} & \multicolumn{1}{c}{R1@0.7} & \multicolumn{1}{c}{mIoU}  \\ \hline
SCDM     & -      & 54.44  & 33.43   & -      & 54.80   & 36.75  & 19.86  & -      & 26.11    & 21.17    & -        & -       \\
2D-TAN   & -      & 42.80  & 23.25   & -      & 58.75  & 44.05  & 27.38  & -      & 35.17    & 25.17    & 11.65    & 24.16     \\
VSLNet   & 64.30  & 47.31  & 30.19   & 45.15  & 63.16  & 43.22  & 26.16  & 43.19  & 29.61    & 24.27    & 20.03    & 24.11           \\
SeqPAN   & 73.84  & 60.86  & 41.34   & 53.92   & 61.65    & 45.50     & 28.37    & 45.11   & 48.64    & 39.64    & 28.07    & 37.17   \\
EMB      & 72.50   & 58.33 & 39.25   & 53.09   & 64.13    & 44.81    & 26.07    & 45.59   & 50.46    & 37.82    & 22.54    & 35.49     \\
EAMAT & 74.19    & 61.69    & 41.96    & 54.45   & 55.33    & 38.07    & 22.87    & 40.12   & 50.11    & 38.16    & 26.82    & 36.43    \\
BAN-APR  &  *74.05 & 63.68    & 42.28    & *54.15 &  *65.11  & 48.12    & 29.67    & *45.87      & 48.24    & 33.74    & *17.44  & *32.95 \\ \hline
CPL   & 66.40     & 49.24    & 22.39    & 43.48   & 55.73    & 31.37    & 12.32    & 36.82   & -  & -  & -  & -   \\
CNM   & 60.04    & 35.15    & 14.95    & - & 55.68    & 33.33    & *12.81    & * 36.15   & -  & -  & -  & -   \\ \hline
FVMR   &- &	55.01&	33.74       &              -    &  60.63 & 	45.00	& 26.85    & - & 41.48	& 29.12	& -       &-             \\ 
CCA   & 70.46    & 54.19    & 35.22    & 50.02   & 61.99    & 46.58    & 29.37    & *45.11 & 45.30     & 32.83    & 18.07    & -   \\ \hline

EMTM (Ours)  & 72.70 & 57.91   &   39.80 &      53.00   &   63.20 &	44.73	& 26.08 & 	45.33  &    45.78 &	34.83 &	23.41 & 34.44     \\ 
$\Delta_{SOTA}$ & \textcolor{red}{$\uparrow$ 2.24}    & \textcolor{red}{$\uparrow$ 2.90}    & \textcolor{red}{$\uparrow$ 4.58}    & \textcolor{red}{$\uparrow$ 2.98}    & \textcolor{red}{$\uparrow$ 1.21}    & \textcolor{blue}{$\downarrow$ 1.85}     & \textcolor{blue}{$\downarrow$ 3.29}     & \textcolor{red}{$\uparrow$ 0.22}     & \textcolor{red}{$\uparrow$ 0.48}  & \textcolor{red}{$\uparrow$ 2.42}  & \textcolor{red}{$\uparrow$ 5.34}  & -  \\
\bottomrule

\end{tabular}
\end{adjustbox}

\caption{ Performance comparison with the state-of-the-art methods. }
\label{table:accuracy}
\vspace{-4mm}
\end{table*}

\section{Experiments}
\vspace{-2mm}

\subsection{Datasets}
\vspace{-2mm}
To evaluate the performance of TSGV, we conduct experiments on three challenging datasets, all the queries in these datasets are in English. Details of these datasets are shown as follows:

  \textbf{Charades-STA}~\cite{gao2017tall} is composed of daily indoor activities videos, which is based on Charades dataset~\cite{sigurdsson2016hollywood}. This dataset contains 6672 videos, 16,128 annotations, and 11,767 moments. The average length of each video is 30 seconds. $12,408$ and $3,720$ moment annotations are labeled for training and testing, respectively;
    
   \textbf{ActivityNet Caption}~\cite{caba2015activitynet} is originally constructed for dense video captioning, which contains about $20$k YouTube videos with an average length of 120 seconds. As a dual task of dense video captioning, TSGV utilizes the sentence description as a query and outputs the temporal boundary of each sentence description.  
    
   \textbf{TACoS}~\cite{regneri2013grounding} is collected from MPII Cooking dataset~\cite{regneri2013grounding}, which has 127 videos with an average length of $286.59$ seconds. 

\subsection{Evaluation Metrics}
\vspace{-2mm}
Following existing video grounding works, we evaluate the performance on two main metrics:

\textbf{mIoU:}
``mIoU" is the average predicted Intersection over Union in all testing samples. The mIoU metric is particularly challenging for short video moments;

\textbf{Recall:}
We adopt ``$\textrm{R@}n, \textrm{IoU}=\mu$'' as the evaluation metrics, following~\cite{gao2017tall}. The ``$\textrm{R@}n, \textrm{IoU}=\mu$'' represents the percentage of language queries having at least one result whose IoU between top-$n$ predictions with ground truth is larger than $\mu$.  In our experiments, we reported the results of $n=1$ and $\mu\in\{0.3, 0.5, 0.7\}$.

\textbf{The Metric of Efficiency}:
Time, FLOPs, and Params are used to measure the efficiency of the model.
Specifically, the time refers to the entire inferring time from the input of  the video and query pair to the output of the prediction. FLOPs refers to floating point operations, which is used to measure the complexity of the model. Params refers to the model parameter size except the word embedding.

\subsection{Implementation Details}\label{implementation}
\vspace{-2mm}
For language query $Q$, we use the $300$-D GloVe~\cite{pennington2014glove} vectors to initialize each lowercase word, which are fixed during training. Following the previous methods, 3D convolutional features (I3D) are extracted to encode videos.
We set the dimension of all the hidden layers as $128$, the kernel size of the convolutional layer as $7$, and the head size of multi-head attention as $8$ in our model. 
For all datasets, models are trained for $100$ epochs. The batch size is set to $16$. The dropout rate is set as 0.2. 
Besides, an early stopping strategy is adopted to prevent overfitting.
The whole framework is trained by Adam optimizer with an initial learning rate of 0.0001. The loss weight $\alpha$ is set as 0.1 in all the datasets. The temperate was set to 1, 3, 3 on Charades-STA, ActivityNet, and TACoS.
The pre-trained teacher models are selected in SeqPAN, BAN-APR, EAMAT, and CCA.
More ablation studies can be found in Section~\ref{sec:ablation}. All experiments are conducted on an NVIDIA RTX A5000 GPU with 24GB memory. All experiments were performed three times, and reporting the average of performance.

\subsection{Comparison with State-of-the-art Methods}
\vspace{-2mm}
We strive to gather the most current approaches, and compare our proposed model with the following state-of-the-art baselines on three benchmark datasets: 

\begin{itemize}
    \item  Proposal-based Methods: SCDM \cite{yuan2019semantic}, 2D-TAN \cite{zhang2020learning}, BAN-APR \cite{dong2022boundary}.
    \item Proposal-free Methods:  VSLNet \cite{zhang2020span},  SeqPAN \cite{zhang2021parallel}, EMB \cite{huang2022video},  EAMAT \cite{yang2022entity}.
    \item Weakly Supervised Methods: CPL \cite{zheng2022weakly},  CNM \cite{zheng2022weakly}
    \item Fast Methods: FVMR \cite{gao2021fast},  CCA \cite{wu2022learning}
\end{itemize}
The best performance is highlighted in \textbf{bold} and the second-best is highlighted with \underline{underline} in tables.

\textbf{Overall Efficiency-Accuracy Analysis} 

Considering that fast TSGV task pays the same attention to efficiency as accuracy, we evaluate FLOPs, Params, and Times for each model. For a fair comparison, the batch size is set to 1 for all methods during inference. Besides, we also calculate the sum of the accuracy in terms of “R1@0.3” and “R1@0.5”, named sumACC to evaluate the whole performance of each model.

As Table \ref{table:efficiency} shows, our method surpasses all other methods and achieves the highest speed, minimal FLOPs and Params on all three datasets. 
We note that EMTM is at least 2000 times fewer in FLOPs than state-of-the-art proposal-based models (SCDM and 2D-TAN). According to sumACC, EMTM outperforms these two models by gains of at most 26.75\% on Charades-STA and 14.30\% on ActivtyNet.
Despite the parameter size of VSLNet is at the same level as our method, we outperform it significantly in terms of accuracy, achieveing 15.30\% absolute improvement by “sumACC” on Charades-STA.
When it comes to CCA, which is proposed for fast TSGV, EMTM outperforms 16950x fewer in FLOPs and 121x fewer in model parameter size on Charades-STA.
The above comparison illustrates that our method has significant efficiency and accuracy advantages.

\textbf{Accuracy Analysis}

As shown in Table \ref{table:accuracy}, we can observe that our method performs better than extensive methods in most metrics on three benchmark datasets. Compared with FVMR and CCA, our model performers better in all metrics. Especially, EMTM achieves an absolute improvement of 4.58\% on Charades- STA and 5.34\% on TACoS on the metric "R@1, IoU=0.7", which is a more crucial criterion with higher quality.

The performance of our model on ActivityNet is slightly lower than CCA. The supposed reason may be that ActivityNet is more challenging since it covers a wide range of videos, not limited to daily indoor activities videos in Charades-STA and cooking videos in TACoS. In such cases, label distillation may not effectively capture the key features of the dataset, resulting in limited performance gains. Besides, the effectiveness of label distillation often depends on the performance of the teacher model. The backbone we use is SeqPAN, which has relatively poor performance on ActivityNet, limiting our upper bound even with label distillation. However, our framework can be adapted to any VMR model. If we replace it with the more powerful upcoming backbone models, the performance will surpass the current version.



\begin{table}[]
\begin{adjustbox}{width=\linewidth}
\renewcommand\arraystretch{1.5}
\label{tabelab1}
\begin{tabular}{l|m{1.1cm}<{\centering}m{1.5cm}<{\centering}|cccc}
\toprule
Method & Shared Encoder & Label Distillation & \makecell*[c]{R1@0.3} & \makecell*[c]{R1@0.5} & \makecell*[c]{R1@0.7} & \makecell*[c]{mIoU} \\ \hline
EMTM w/o SE-LD &  \texttimes               & \texttimes   & $70.19_{\textcolor{blue} {-0.99}}^{\textcolor{red}{+0.97}} $  & $ 56.23_{\textcolor{blue}{-1.01}}^{\textcolor{red}{+0.62}}  $  & $ 36.49_{\textcolor{blue}{-0.74}}^{\textcolor{red}{+0.39}} $    & $  51.34_{\textcolor{blue}{-1.06}}^{\textcolor{red}{+0.98}} $  \\
EMTM w/o SE &\texttimes              & \checkmark  & $  \textbf{73.33}_{\textcolor{blue}{-1.34}}^{\textcolor{red}{+0.84}} $  & $  \textbf{58.05}_{\textcolor{blue}{-0.25}}^{\textcolor{red}{+0.26}}  $  & $  \underline{38.36}_{\textcolor{blue}{-0.21}}^{\textcolor{red}{+0.17}} $  & $ \textbf{53.31}_{\textcolor{blue}{-0.91}}^{\textcolor{red}{+0.54}} $  \\
EMTM w/o LD &\checkmark              & \texttimes   & $ 72.62_{\textcolor{blue}{-0.52}}^{\textcolor{red}{+0.69}}   $  & $ 56.51_{\textcolor{blue}{-0.84}}^{\textcolor{red}{+1.18}}  $   & $ 37.54_{\textcolor{blue}{-0.50}}^{\textcolor{red}{+0.85}}  $   & $ 52.39_{\textcolor{blue}{-0.37}}^{\textcolor{red}{+0.60}} $  \\
EMTM &\checkmark              & \checkmark   & $  \underline{72.70}_{\textcolor{blue}{-0.55}}^{\textcolor{red}{+0.47}}  $   & $ \underline{57.91}_{\textcolor{blue}{-0.65}}^{\textcolor{red}{+0.75}} $    & $ \textbf{39.80}_{\textcolor{blue}{-0.12}}^{\textcolor{red}{+0.12}}  $   & $ \underline{53.00}_{\textcolor{blue}{-0.33}}^{\textcolor{red}{+0.21}} $  \\ \bottomrule
\end{tabular}
\end{adjustbox}
\vspace{-2mm}
\caption{Effects of Components on Charades-STA.}
\label{tabel:component_charades}
\vspace{-4mm}
\end{table}

\begin{table}[htbp]
\begin{adjustbox}{width=\linewidth}
\renewcommand\arraystretch{1.5}
\label{tabelab2}
\begin{tabular}{l|m{1.1cm}<{\centering}m{1.5cm}<{\centering}|cccc}
\toprule
Method & Shared Encoder & Label Distillation & \makecell*[c]{R1@0.3} & \makecell*[c]{R1@0.5} & \makecell*[c]{R1@0.7} & \makecell*[c]{mIoU} \\ \hline
EMTM w/o SE-LD &\texttimes              & \texttimes   & $ 62.06_{\textcolor{blue}{-0.85}}^{\textcolor{red}{+0.99}}  $    & $ 43.90_{\textcolor{blue}{-0.39}}^{\textcolor{red}{+0.25}} $    & $ 25.63_{\textcolor{blue}{-0.13}}^{\textcolor{red}{+0.07}}  $   & $ 44.52_{\textcolor{blue}{-0.38}}^{\textcolor{red}{+0.55}} $  \\
EMTM w/o SE &\texttimes              & \checkmark  & $  \underline{63.19}_{\textcolor{blue}{-0.22}}^{\textcolor{red}{+0.35}} $  &$  44.11_{\textcolor{blue}{-0.27}}^{\textcolor{red}{+0.26}}$  & $ 25.74_{\textcolor{blue}{-0.32}}^{\textcolor{red}{+0.41}} $ & $ 45.15_{\textcolor{blue}{-0.03}}^{\textcolor{red}{+0.03}} $  \\
EMTM w/o LD &\checkmark              & \texttimes   &$  62.98_{\textcolor{blue}{-0.40}}^{\textcolor{red}{+0.33}} $    & $ \textbf{44.68}_{\textcolor{blue}{-0.15}}^{\textcolor{red}{+0.19}} $    & $ \textbf{26.10}_{\textcolor{blue}{-0.06}}^{\textcolor{red}{+0.12}}$     & $ \underline{45.22}_{\textcolor{blue}{-0.12}}^{\textcolor{red}{+0.11}} $  \\
EMTM &\checkmark              & \checkmark   & $ \textbf{63.20}_{\textcolor{blue}{-0.58}}^{\textcolor{red}{+0.30}}    $ &$  \underline{44.73}_{\textcolor{blue}{-0.33}}^{\textcolor{red}{+0.58}} $    &$  \underline{26.08}_{\textcolor{blue}{-0.31}}^{\textcolor{red}{+0.27}}  $   & $ \textbf{45.33}_{\textcolor{blue}{-0.31}}^{\textcolor{red}{+0.19}} $  \\ \bottomrule
\end{tabular}
\end{adjustbox}
\vspace{-1mm}
\caption{Effects of Components on ActivityNet}
\label{tabel:component_anet}
\vspace{-5mm}
\end{table}

\begin{table}[htbp]
\footnotesize
\begin{adjustbox}{width=0.9\linewidth}
\renewcommand\arraystretch{1.5}
\label{tabelab3}
\begin{tabular}{l|m{1.1cm}<{\centering}|m{1.5cm}<{\centering}|c|c}
\toprule
Method & \makecell*[c]{R1@0.3} & \makecell*[c]{R1@0.5} & \makecell*[c]{R1@0.7} & \makecell*[c]{mIoU} \\ \hline
$\text{EMTM}_{w/o KAU}$  & 72.47	& 57.63	 & 38.58 & 52.18  \\
$\text{EMTM}_{3,3,3}$   & 71.08	& 56.99	 & 37.82 & 52.06  \\
$\text{EMTM}$           & 72.70	& 57.91	 & 39.80	 & 53.00  \\
 \bottomrule
\end{tabular}
\end{adjustbox}
\vspace{-2mm}
\caption{Effects of KAU on Charades-STA.}
\label{tabel:kau_charades}
\vspace{-5mm}
\end{table}

\subsection{Ablation Studies}\label{sec:ablation}
\vspace{-2mm}
In this part, we perform ablation studies to analyze the effectiveness of the EMTM. All experiments are performed three times with different random seeds.

\vspace{-2mm}
\subsubsection{Effects of Components} \label{sec:main_components}
\vspace{-2mm}

In our proposed framework, we design the sharing encoders (SE) to learn shallow knowledge from the teacher by label distillation(LD). To better reflect the effects of these two main components, we measure the performance of different combinations. 
As table \ref{tabel:component_charades}  and \ref{tabel:component_anet} show, each interaction component has a positive effect on the TSGV task. On Charades-STA, the full model outperforms \textbf{w/o SE} by gains of 1.44\% on metrics “R@1, IoU={0.7}” and exceeds “w/o LD” by 
on the all metrics. Besides, the full model also outperforms “w/o SE-LD” by a large margin on all metrics. Similarly, our full model has made significant improvements in all metrics compared with the variant "EMTM w/o SE-LD" on ActivityNet.


Additionally, we conduct two ablation experiments of KAU for analysis in table \ref{tabel:kau_charades}. As shown, KAU with the kernel size 3, 5, 7 instead of 3, 3, 3 does work. 

\vspace{-3mm}
\subsubsection{Effect of Number of Teacher Models}
\vspace{-2mm}

\begin{figure}[htbp]   
  \begin{minipage}[t]{0.5\columnwidth} 
    \centering   
    \includegraphics[width=\linewidth]{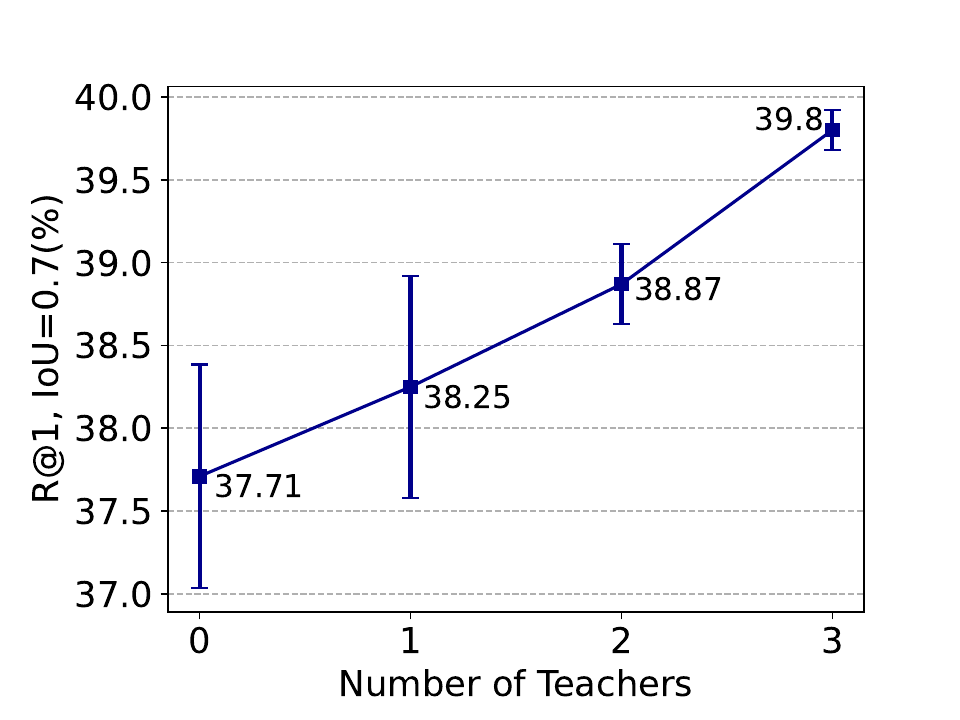}     
    \label{fig:side:a}   
  \end{minipage}%
  \begin{minipage}[t]{0.5\columnwidth}   
    \centering   
    \includegraphics[width=\linewidth]{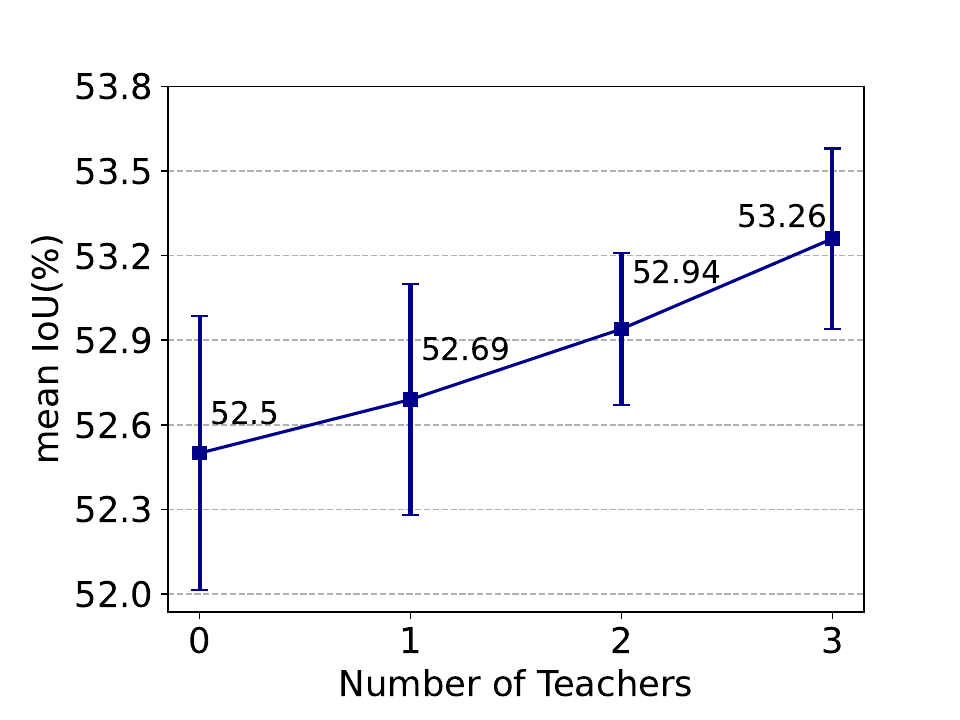}    
    \label{fig:side:b}   
  \end{minipage}   
\vspace{-5mm}
\caption{Effect of the Number of Teacher Models on Charades-STA. In detail, we adopt EAMAT, EAMAT \& BAN-APR, EAMAT \& BAN-APR \& SeqPAN, which correspond to one teacher, two teachers and three teachers respectively.}
\label{teacher_number}
\vspace{-4mm}
\end{figure}

We investigate the influence of different numbers of teacher models on Charades-STA. As shown in Figure \ref{teacher_number}, the performance presents a rising tendency with the increase of teachers. 
According to the results, we realize that our improvements are not only from soft targets with one single teacher, but also from the learning of structural knowledge and intermediate-level knowledge with fused multi-teacher teaching. Multiple teachers make knowledge distillation more flexible, while ensemble helps improve the training of students and transfer related information of examples to them.

\begin{figure}[htbp]
    \centering
    \includegraphics[width=\linewidth]{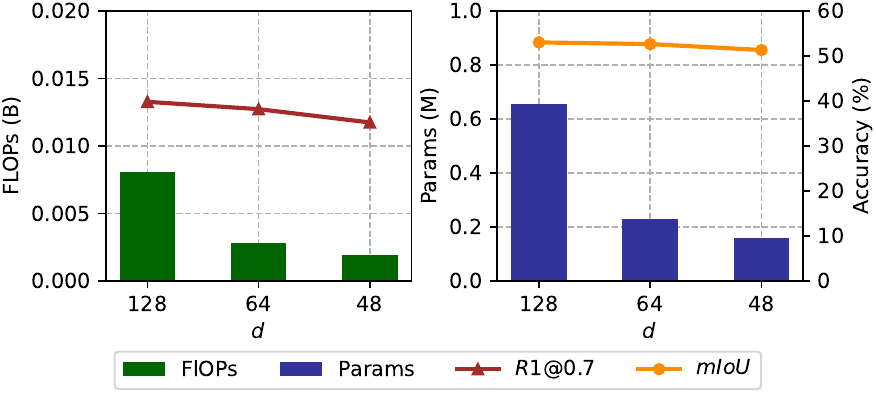}
    \caption{Effect of different degrees of lightweight by adjusting the hidden dimension $d$.}
    \label{fig:dimension}
    \vspace{-3mm}
\end{figure}

\begin{figure}[htbp]
    \centering
    \includegraphics[width=0.98\linewidth]{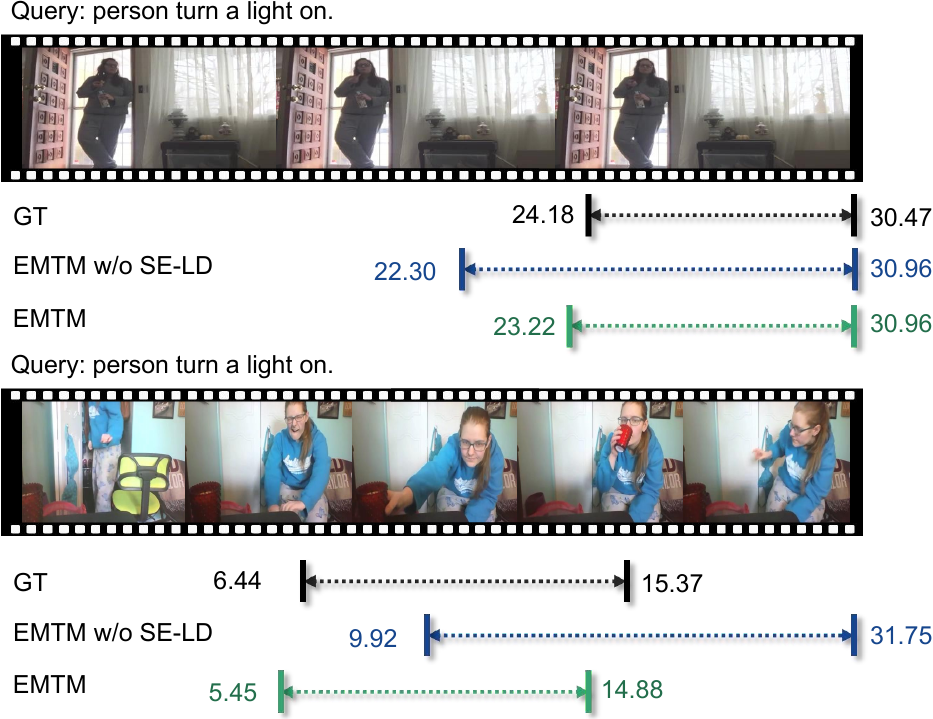}
    \caption{Examples of visualization of EMTM w/o SE-LD and EMTM on the Charades-STA. }
    \label{fig:visual}
    \vspace{-3mm}
\end{figure}

\vspace{-3mm}
\subsubsection{Effect of Different Degree of Lightweight Models}
\vspace{-2mm}

We evaluate the influence of different degrees of lightweight models by adjusting their hidden dimension $d$ on Charades-STA. As shown in Figure \ref{fig:dimension}, obviously as $d$ decreases, the FLOPs and model parameter size will decline, which would also reduce the performance. From 128 to 64 for $d$, both R1@0.7 and mIoU reduce by about 5\%, while FLOPs and model parameter size drop by a small margin. For the trade-offs, we select 128 as the hidden dimension.

\subsection{Qualitative Analysis}
\vspace{-2mm}
Two samples of prediction on Charades-STA are depicted in Figure \ref{fig:visual}.
The first sample indicates our approach can refine the predictions when the basic model already obtained satisfactory results.
The second sample shows the basic model tends to predict the boundary position, possibly due to its limited understanding of the video.
As a result, the model relies on biased positional information to make moment predictions. 
However, utilizing a shared encoder and label distillation approach can provide additional information that enables the model to more precisely predict the moment boundary.

\section{Conclusion}
\vspace{-2mm}
In this paper, we focus on the efficiency of the model on TSVG and try to expand the efficiency interval to cover the entire model.
 A knowledge distillation framework (EMTM) is proposed, which utilizes label distillation from multiple teachers and a shared encoder strategy. 
In the future, we will pay attention to video feature extraction in TSGV, which is also a time-consume process.
We will propose an end-to-end model that input video frames.


{
    \small
    \bibliographystyle{ieeenat_fullname}
    \bibliography{main}
}

\end{document}